# Benchmark datasets driving artificial intelligence development fail to capture the needs of medical professionals

Kathrin Blagec, M.D.[1] *, Jakob Kraiger, M.D.[1] *, Wolfgang Frühwirt, PhD [2],
Matthias Samwald, PhD [1]

1) Institute of Artificial Intelligence, Medical University of Vienna, Vienna, Austria

2) Machine Learning Research Group, University of Oxford, Oxford, UK

* Have contributed equally

Corresponding author:

Assoc.Prof. Matthias Samwald, PhD
Institute of Artificial Intelligence
Medical University of Vienna
Währingerstraße 25a, BT1, OG1
1090 Vienna, Austria
matthias.samwald [at] meduniwien.ac.at

# Abstract

Publicly accessible benchmarks that allow for assessing and comparing model performances are important drivers of progress in artificial intelligence (AI). While recent advances in AI capabilities hold the potential to transform medical practice by assisting and augmenting the cognitive processes of healthcare professionals, the coverage of clinically relevant tasks by AI benchmarks is largely unclear. Furthermore, there is a lack of systematized meta-information that allows clinical AI researchers to quickly determine accessibility, scope, content and other characteristics of datasets and benchmark datasets relevant to the clinical domain.

To address these issues, we curated and released a comprehensive catalogue of datasets and benchmarks pertaining to the broad domain of clinical and biomedical natural language processing (NLP), based on a systematic review of literature and online resources. A total of 450 NLP datasets were manually systematized and annotated with rich metadata, such as targeted tasks, clinical applicability, data types, performance metrics, accessibility and licensing information, and availability of data splits. We then compared tasks covered by AI benchmark datasets with relevant tasks that medical practitioners reported as highly desirable targets for automation in a previous empirical study.

Our analysis indicates that AI benchmarks of direct clinical relevance are scarce and fail to cover most work activities that clinicians want to see addressed. In particular, tasks associated with routine documentation and patient data administration workflows are not represented despite significant associated workloads. Thus, currently available AI benchmarks are improperly aligned with desired targets for AI automation in clinical settings, and novel benchmarks should be created to fill these gaps.

# 1. Introduction

Recent advances in artificial intelligence (AI) capabilities hold the potential to transform medical practice. In contrast to the oft-discussed replacement of medical professionals, fields of application in the foreseeable future will rather lie in assisting and augmenting the cognitive processes of professionals.

Recent years have witnessed a rapid increase in research on algorithms that are aimed at a wide variety of clinical tasks, such as diabetic retinopathy detection from fundus images, visual question-answering from radiology images, or information synthesis and retrieval. [1–5] Besides tasks aimed directly at clinical care, other more administrative healthcare tasks that may help to reduce clinicians' administrative workloads—such as automatic ICD coding from electronic health record (EHR) data—have also been targeted. [6–8]

As AI algorithms and training paradigms become better and more versatile, the availability of high-quality, representative data to train and validate machine learning models is key to optimally harnessing their potential for biomedical research and clinical decision making.

In AI research, model performance is commonly evaluated and compared using benchmarks. A benchmark constitutes a task (e.g., named entity recognition), a dataset representative for the task and one or more metrics to evaluate model performance (e.g., F1 Score). Benchmark datasets are commonly publicly available to researchers and therefore provide a transparent, standardised way to assess and compare model performance. Consequently, benchmarks also act as drivers for AI development, and a rapidly growing body of research is devoted to understanding, critically reflecting and improving AI benchmarking and capability measurement [9–14].

While benchmarks exist for a large number of general domain AI tasks, the coverage of biomedical and clinical tasks is largely unclear. Having a clear understanding of what kind of tasks the biomedical and clinical AI research community is currently tackling, which of these tasks are currently covered by benchmarks and which are not, helps to inform the future creation of benchmarks.

A systematic model of biomedical and clinical tasks can further offer valuable insights into how different research focus areas might synergize and allow insights for future research focus areas. Currently, there is a lack of systematized meta-information on biomedical and clinical datasets and benchmarks, that allow researchers to quickly determine their accessibility, scope, content and other characteristics. [15]

Several initiatives to index datasets and benchmark datasets relevant to AI research have been introduced in recent years. These include the 'Papers with Code' datasets[1], 'Hugging Face' datasets[2] or the Online Registry of Biomedical Informatics Tools (ORBIT) Project'.[3] [16] These are, however, either not maintained anymore (ORBIT), focused on making datasets available for use in programming frameworks (Hugging Face) or capturing benchmark results (Papers with Code) rather than on systematizing and modeling datasets and tasks. Furthermore, the latter two are focused on general domain tasks and datasets, with biomedical and clinical datasets making up only a small fraction of the current databases.

Finally, the future utility of AI for healthcare therefore hinges on how well AI benchmarks reflect the actual needs in healthcare. To the best of our knowledge, currently there exists no study that investigates this essential question.

This paper aims to address these issues in a threefold way, focused on natural language processing (NLP) tasks:

- First, we introduce a comprehensive curated catalogue of 450 biomedical and clinical NLP datasets and benchmark datasets based on a systematic literature review covering both biomedical and computer science literature and grey literature data sources.
- Second, we manually systematize and annotate these datasets and benchmarks with meta-information, such as accessibility, performance metrics, availability of data splits and associated tasks, while considering interoperability and harmonisation with existing ontologies, such as EDAM, SNOMED CT and ITO. [17,18]
- Finally, based on this data source, we analyze the current availability of clinically relevant AI benchmarks, their overlap with actual needs in healthcare, their strengths and weaknesses, and identify opportunities for advancing AI for clinical applications.

[1] https://paperswithcode.com/datasets
[2] https://huggingface.co/docs/datasets/
[3] https://wayback.archive-it.org/org-350/20200626193002/https:/orbit.nlm.nih.gov/

# 2. Materials and methods

## 2.1. Compilation of dataset catalogue

### 2.1.1 Data sources

PubMed (MEDLINE)[4] and arXiv[5] were selected as the main sources to ensure coverage of both biomedical and computer science literature. The MEDLINE database contains bibliographic information on more than 26 million biomedical journal articles as of January 2020. arXiv is an open access pre-print server for scientific papers covering a wide range of fields, including computer science. We considered using additional search engines such as Semantic Scholar for the review, but decided to fall back on PubMed and arXiv due to e.g., missing Boolean search functions or intransparent search modalities as described previously by Gusenbauer and Haddaway. [19]

In addition to the systematic literature review,  we included sources such as NLP and ML challenge and shared-task websites, such as BioASQ[6] and n2c2[7].

### 2.1.2 Literature review

We focused our review on benchmarks for NLP tasks. PubMed was queried using the web interface and results as of November 12, 2020 were exported. arXiv was queried using its API on November 27, 2020. Records were converted from XML to CSV. Further screening and annotation of PubMed and arXiv records was done via Google sheets between December 2020 and February 2021.

The literature review flowchart and queries used to obtain the results are shown in Figure 1. Publication abstracts and full texts were further screened for the mention of datasets and dataset names and associated tasks were extracted.

[4] https://www.nlm.nih.gov/bsd/medline.html
[5] https://arxiv.org
[6] http://www.bioasq.org/
[7] https://portal.dbmi.hms.harvard.edu/

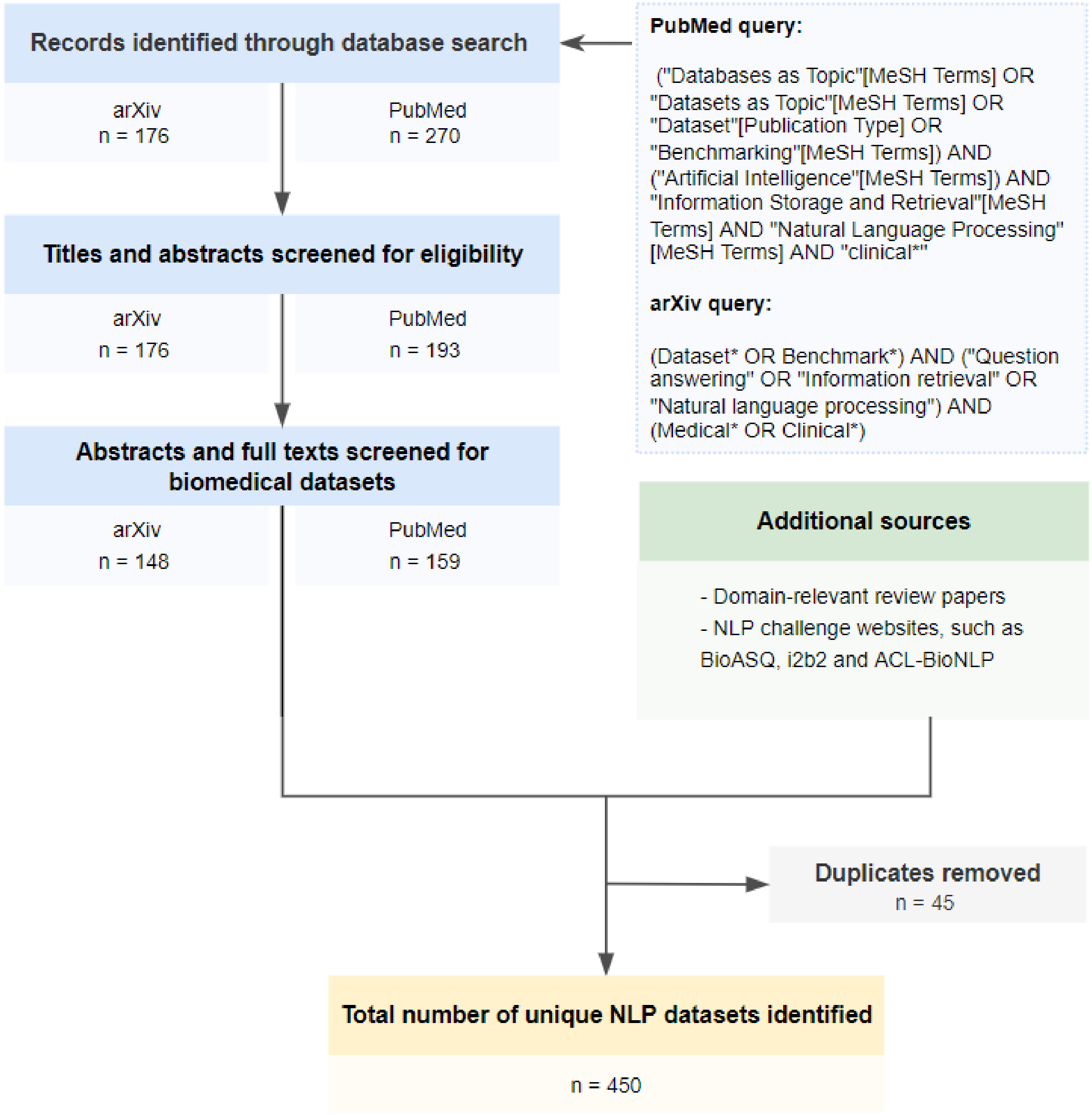

**Figure 1. Literature review flowchart, queries and other data sources.**

### 2.1.3 Annotation process

Extracted datasets were de-duplicated and annotated for their meta-information. In many cases datasets appeared under different naming variations or did not have an explicit name. We normalized the dataset names to the best of our knowledge with, e.g., names from the official dataset websites or repositories as a reference for the normalised names. For datasets with no explicit names we instead used a surrogate name in the form of a short description as reported in the respective paper. For each dataset, alias names occurring in the identified records were annotated.

**Table 1. Definitions of terms.**

| Term | Definition |
|---|---|
| Clinical | “Relating to the examination and treatment of patients and their illnesses” (Oxford dictionary) |
| Benchmark dataset | “Any resource that has been published explicitly as a dataset that can be used for evaluation, is publicly available or accessible on request, and has clear evaluation methods defined” |
| Clinically relevant benchmark dataset | “Benchmark datasets directly relating to the entirety of processes involved in the examination and treatment of patients and their illnesses.” |
| Information retrieval (IR) | “Obtaining information system resources that are relevant to an information need from a collection of those resources” (Source: Wikipedia). |
| Question answering (QA) | “Building systems that automatically answer questions posed in a natural language” (Source: Adapted from Wikipedia). |
| Clinical care task | Tasks that are directly related to the examination and treatment of patients and their illnesses. Includes reviewing and searching for medical information using a variety of information sources, such as books, scientific literature or web-based information content. Includes the analysis and interpretation of diagnostic tests including medical imaging results. |
| Administrative task | Administrative tasks include, e.g., scheduling and managing patient appointments, filing, updating, and organizing patient records or coding medical records for billing. |
| Scientific task | Tasks related to the coordination, conduct or reporting of clinical scientific research. Includes e.g., the selection of eligible patients for clinical trials. |

### 2.1.4 Benchmark datasets

We distinguished between datasets and benchmark datasets. We defined benchmark datasets as “any resource that has been published explicitly as a dataset that can be used for evaluation, is publicly available or accessible on request, and has clear evaluation methods defined” (see Table 1). While our analysis is focused on benchmark datasets, we also documented relevant non-benchmark datasets in the catalogue because they may be used for the creation of novel datasets as well as unsupervised pre-training of machine learning models.

### 2.1.5. Tasks

Task descriptions extracted from the identified records were standardized and mapped to broad task families based on two relevant ontologies, i.e. the Intelligence Task Ontology (ITO)[8] and SNOMED CT[9]. [20]

### 2.1.6. Clinical relevance

We defined clinical tasks as tasks relating to the entirety of processes involved in the examination and treatment of patients and their illnesses (working definition adapted from Oxford Dictionary). Clinical work activities often additionally include administrative tasks, such as 'ICD coding' for billing. We therefore included such tasks in our definition of clinical tasks, and made a distinction between ‘clinical care’ tasks and ‘administrative’ tasks only later in our analysis (see Section ‘Comparison with healthcare practitioners’ preferences for task automation’).

We further distinguished between *directly* and *indirectly* clinically relevant tasks, where directly relevant tasks represent actual clinical tasks (e.g., answering clinical questions), and indirectly relevant tasks refer to tasks that may be relevant to improve clinical tasks but do not directly represent such tasks (e.g., information extraction for clinical documents).

Furthermore, we classified clinically relevant tasks based on their main target group, e.g. clinicians, radiologists, pathologists or patients. Annotations on clinical relevance and target groups (e.g., clinicians, radiologists, patients) were done by an MD (JK) and an MD with experience in NLP (KB). Any discrepancies or ambiguities in the annotations were discussed between all authors.

### 2.1.7. Other annotations

Other annotated meta-information included type of the source data used to create the dataset (e.g., clinical notes), dataset accessibility, licensing type, language, links to source publications and, if available, links to dataset repositories. For benchmark datasets, we additionally annotated performance metrics and the availability of data splits.

[8] https://bioportal.bioontology.org/ontologies/ITO
[9] https://www.snomed.org/

The complete list of annotated fields including their descriptions can be found in Table 2.

**Table 2. Description of annotation fields.**

| Field | Annotation | Description |
|---|---|---|
| Name (or description if no name available) | | Name of the dataset or description if no official name is available |
| Task | | Task as described in the source paper (if available). |
| Mapped task | | Mapped task from relevant ontologies, e.g., Intelligence Task Ontology (ITO) and SNOMED CT. |
| Id(s) of mapped task | | Id(s) of ontology classes representing the mapped task. |
| Data basis | | Data type of the source data used to create the dataset, e.g., “Clinical notes / EHR data”. |
| Availability of evaluation criteria | ‘Yes’ | The paper or original source describes evaluation criteria that are used to assess a model’s performance on the respective task. |
| | ‘No’ | The dataset was published without any specific evaluation criteria. |
| Metrics | | Performance metrics for evaluation on benchmark datasets, e.g., F-Measure. |
| Data splits | ‘Official’ | The dataset can be downloaded together with an official train / validation / test split. |
| | ‘Described’ | A data split is described in the source paper but not available together with the data. |
| | ‘Not described’ | The source paper does not contain any information on data splits. |
| | ‘Not available’ | There is no data split. |
| Accessibility | ‘Public’ | The dataset is publicly available and can be accessed and downloaded by everyone. |
| | ‘Public (planned)’ | The dataset is not yet publicly available but there are plans to make it available. |
| | ‘Upon registration’ | The dataset can be accessed and downloaded after filling out a registration form. |
| | ‘On request’ | The dataset can be requested from the owner(s) via email. |
| | ‘Unknown’ | The availability of the dataset is unknown, i.e. it is not made available and there is no explicit statement about its availability. |
| | ‘Not available’ | It is explicitly stated that the dataset is not available due to, e.g., privacy or other reasons. |
| Data license | | Data license type, e.g., Creative Commons 4.0 BY. |
| Data license notes | | Additional notes on data license or data use requirements. |
| Clinical relevance | 'Directly' | Tasks relating to the entirety of processes involved in the examination and treatment of patients and their illnesses (Working definition adapted from Oxford Dictionary). |
| | 'Indirectly' | Tasks that may be relevant to improve clinical tasks but do not directly represent such tasks |
| | 'Not relevant' | Tasks not relevant for clinical applications. |
| Relevance for IR / QA | 'Directly' | Tasks that directly represent information retrieval (IR) or question answering (QA) tasks, e.g., literature queries or question answering tasks. |
| | 'Indirectly' | Tasks that may be relevant to IR and QA tasks but do not directly represent such tasks, e.g., Named Entity Recognition or Abbreviation expansion. |
| | 'Not relevant' | Tasks not relevant for IR and QA. |
| Primary target group | 'Patients' | The task represents a task primarily relevant to patients, e.g., searching health information on the web. |
| | 'Clinicians' | The task represents a task relevant to clinicians, e.g., retrieval of information on contraindications of a drug. |
| | 'Radiologists' | The task represents a task relevant to radiologists, e.g., generating radiology reports. |
| | 'Pathologists' | The task represents a task relevant to pathologists, e.g., visual question answering on pathology images. |
| Link: Source publication(s) | | Link to the source publication(s) the dataset was extracted from. |
| Link: Main reference | | Link to the main publication(s) that describes the dataset (if available). |
| Link: Dataset | | Link to the dataset repository or official website of the dataset. |
| First time published | | Year of first release of the dataset. |
| Language | | Language of the dataset |
| Comment | | Annotator comments |
| Aliases | | Alternative names of the dataset |

## 2.2. Analysis

### 2.2.1. Descriptive statistics of the curated catalogue

We calculated descriptive statistics of the resulting catalogue, such as the number of datasets that qualify as benchmark datasets based on our predefined criteria, accessibility of datasets, availability of data splits, source data types (e.g., clinical notes) and task families (e.g., information extraction).

### 2.2.2. Comparison with healthcare practitioners' preferences for task automation

To compare healthcare practitioners' preferences for automation of their tasks with tasks covered by benchmark datasets, we combined our results with recent work by Frühwirt and Duckworth, which represents the first comprehensive quantitative evidence of healthcare practitioners' preferences regarding the automation of their own work activities based on the O*NET[10] classification scheme. [21,22] The O*NET system is a database of occupational characteristics in terms of required knowledge, skills, tasks and work activities maintained by the U.S. Department of Labor. [21]

For our analysis, we filtered the O*NET database list of work activities for tasks commonly performed by MDs, whereas we excluded tasks primarily done by other personnel involved in healthcare, such as psychologists or technical assistants. Further, only tasks commonly performed by MDs that could potentially be assisted by natural language processing were considered in our analysis. The complete list of work activities is available from the online supplementary material (see Section 'Code and data availability').

To assess task coverage, we then mapped the list of clinical tasks to the list of clinical and biomedical NLP tasks extracted from our dataset of biomedical and clinical datasets and benchmark datasets by assessing whether the NLP task was directly relevant for the clinical task. For example, the NLP task 'visual question answering' was deemed relevant for the clinical tasks 'Process x-rays or other medical images' and 'Analyze test data or images to inform diagnosis or treatment', while the NLP task

---

[10] https://www.onetonline.org/

‘Information retrieval’ was deemed relevant for the clinical task ‘Review professional literature to maintain professional knowledge’.

We further organised the list of clinical tasks into the categories (1) clinical care tasks (e.g., diagnosing a disease) , (2) administrative tasks (e.g., report writing), and (3) scientific tasks (e.g., patient cohort stratification for clinical trials). Tasks occurring in the scientific literature that were targeted at basic clinical/biomedical NLP research, such as ‘Semantic similarity estimation’ or ‘Question entailment detection’ were assigned to a separate class ‘Improving clinical/biomedical NLP’.

# 3. Results

## 3.1. Catalogue of datasets

All analyses are based on version 0.1.3 of the dataset. The catalogue currently contains 450 unique NLP datasets, of which 205 (45.5%) qualify as benchmark datasets based on our criteria (i.e. the availability of evaluation criteria and public accessibility of the dataset or accessibility upon registration or request).

The dataset is released in two formats: As a Google sheet[11] and as a versioned TSV file at Zenodo[12]. [23] Additionally, we make the raw exports of the literature review results available via Zenodo.

Figure 2 shows the characteristics of all datasets (i.e. benchmark datasets and non-benchmark datasets) included in the catalogue in terms of source data, task family, accessibility and clinical relevance.

At the time of analysis, only 28.2% of all datasets were publicly available while 9.5% and 13.1% of datasets were available after undergoing a registration procedure or upon written request, respectively (see Figure 2c). It has, however, to be noted that in many of these cases the applicant may be required to hold a certain position, e.g., verifiably be employed as a researcher at an academic institution.

[11] https://docs.google.com/spreadsheets/d/1QjUxxnZ3tuyW5dj6nkt_o5yJcWUZec4ttfJxO8Zlty4/

[12] DOI: 10.5281/zenodo.4647823

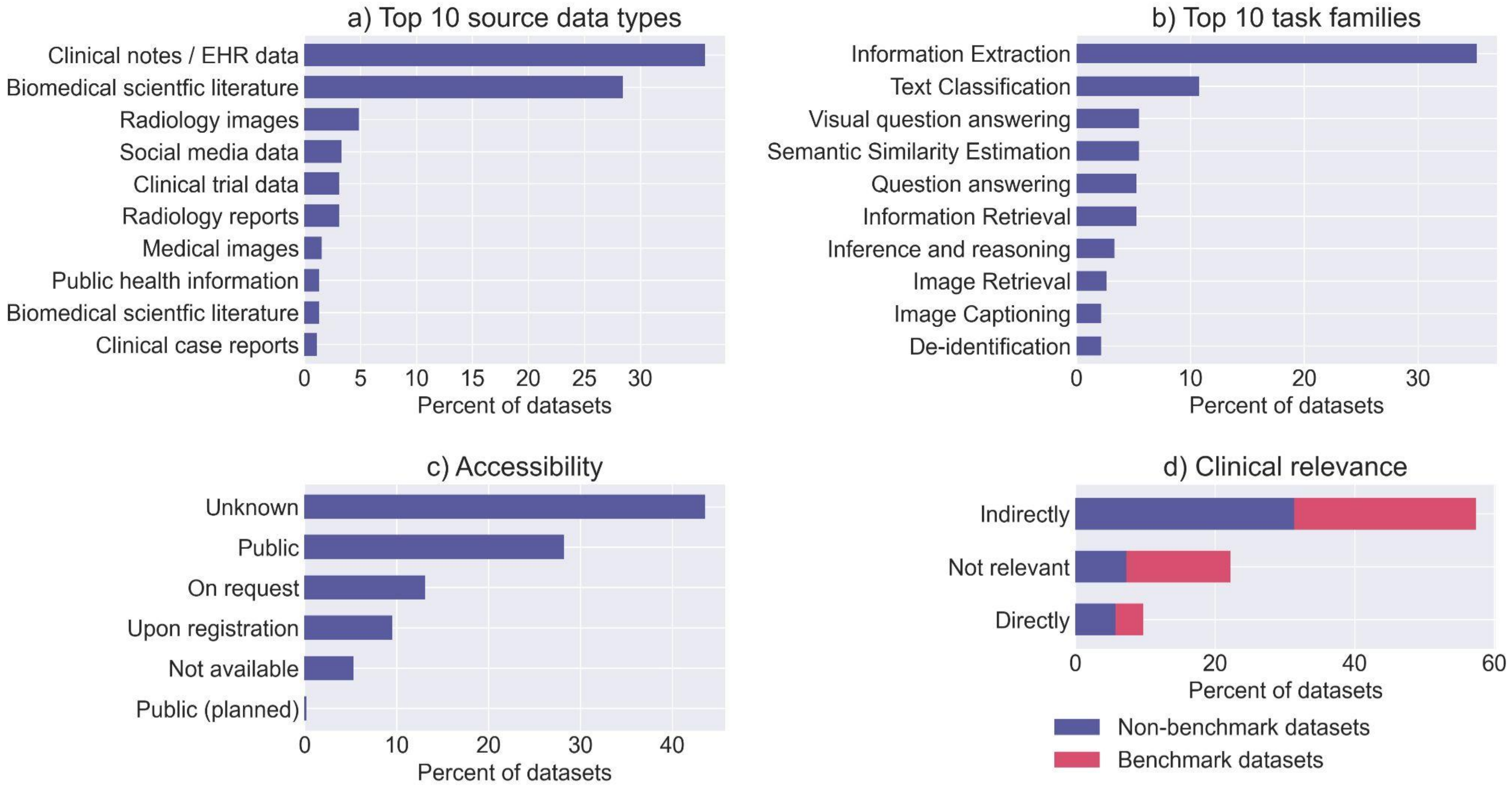

**Figure 2. Characteristics of all datasets (i.e. benchmark datasets and non-benchmark datasets) included in the catalogue.** a) Top 10 source data types. Note: 'Radiology images' and 'Medical images' are included in the list of source data because they are part of the source data of visual question answering datasets. b) Top 10 task families. Only datasets associated with a concrete task as stated in the reference source are included in this chart (n=314). c) Dataset accessibility. d) Proportion of benchmark datasets among directly and indirectly clinically relevant datasets. Only datasets associated with a concrete task as stated in the reference source are included in this chart (n=314).

## 3.2. Analysis of benchmark datasets

The overwhelming majority (86.7%, n=117) of currently available clinically relevant benchmarks are focused on narrow technical aspects of clinical and biomedical AI tasks. In contrast, few benchmarks with direct clinical relevance exist (13.3%, n=18).

Figure 3 shows characteristics of the identified benchmark datasets in terms of source data, task family and the availability of pre-defined data splits.

Benchmarks classified as directly clinically relevant belonged to four broad task groups, i.e., visual question-answering in the areas of radiology and histopathology (n=6), text-based information retrieval and question-answering in general clinical domains (n=7), radiological/histopathological report generation (n=3), and radiological image annotation (n=2). Table 3 lists the identified benchmarks with direct clinical relevance.

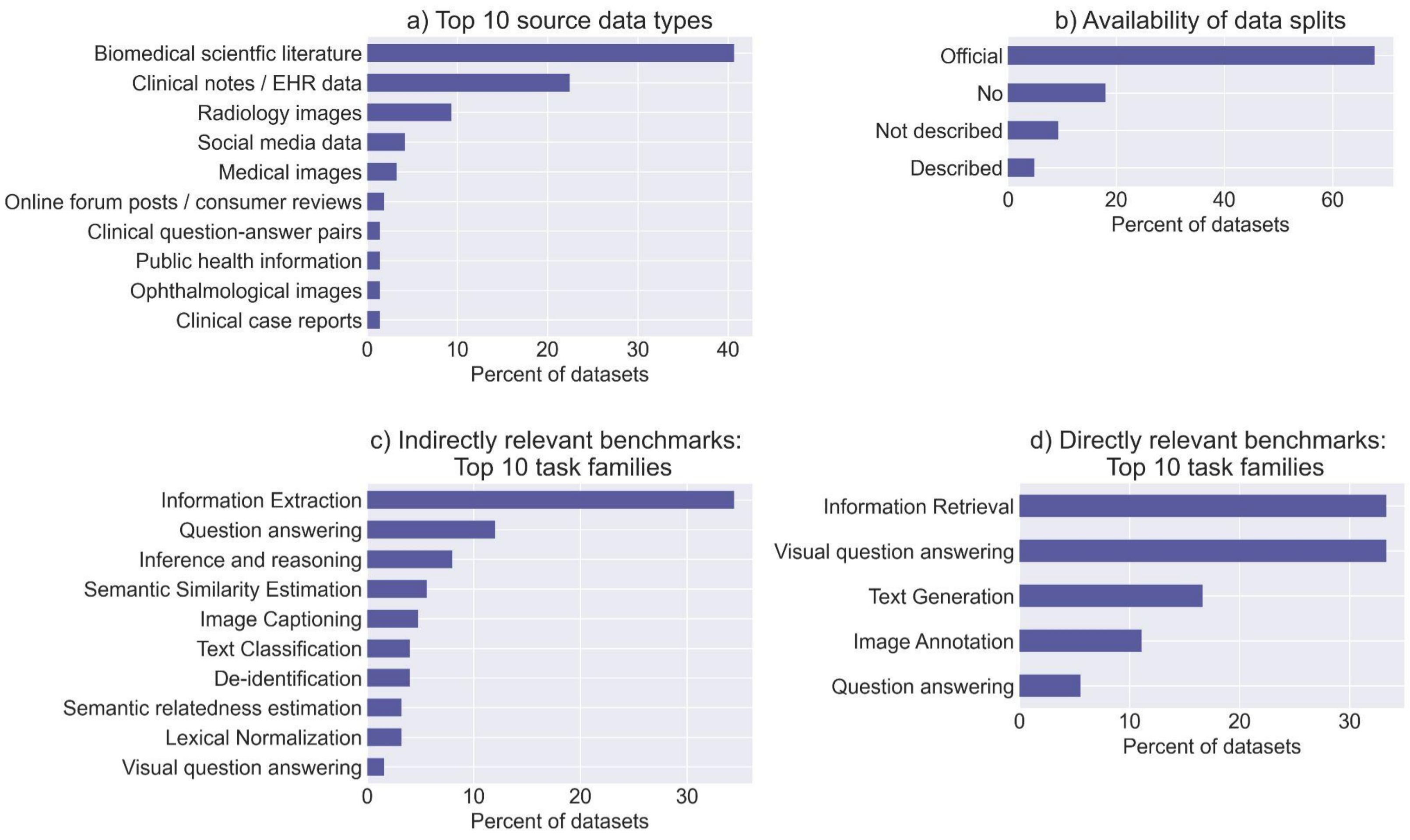

**Figure 3. Characteristics of identified benchmark datasets.** a) Top 10 source data types. Note: 'Radiology images' and 'Ophthalmological images' are included in the list of source data because they are part of the source data of visual question answering datasets. b) Availability of data splits. c) Top 10 task families of indirectly relevant benchmarks (n=117). d) Top 10 task families of directly relevant benchmarks (n=18).

**Table 3. Identified benchmarks with direct clinical relevance and dataset specifications.** Q: Question, A: Answer. *: Example was abbreviated from original. Benchmarks that are published on a regular basis as part of recurring competitions and thus share task descriptions and/or have similar datasets were collated into single rows.

| **Benchmark name** | **Task description** | **Primary target groups** | **Example item** | **Accessibility** |
|---|---|---|---|---|
| TREC CDS task 2014, 2015 and 2016 [24] | Information retrieval for evidence-based full-text literature to support diagnosis, treatment, and testing decisions. | Clinicians | Topic: "A woman in her mid-30s presented with dyspnea and hemoptysis. CT scan revealed a cystic mass in the right lower lobe. Before she received treatment, she developed right arm weakness and aphasia. She was treated, but four years later suffered another stroke. Follow-up CT scan showed multiple new cystic lesions."* | Public |
| TREC-COVID [25] | Information retrieval for evidence-based full-text literature on COVID-19 to support diagnosis, treatment, and testing decisions. | Clinicians | Query: "COVID-19 cytokine storm"<br>Question: "What is the mechanism of cytokine storm syndrome on the COVID-19?" | Public |
| Dataset based on 56 real life diagnostic cases (FindZebra) [26] | Information retrieval for information on rare diseases. | Clinicians | Query: "25 year old, woman, conjunctival hyperaemia, interstitial keratitis, moderate bilateral sensorineural hearing loss, tinnitus, dizziness, nausea and vertigo"<br>Diagnosis: "Cogan's syndrome" | Public |
| emrQA [27] | Question-answering on electronic medical records. | Clinicians | Q: "Why did the patient need lantus?"<br>A: "Blood sugar was 308 at 6 pm last night, 340 at 7 am this morning."* | Upon registration |
| ShAReCLEF eHealth 2013 Task 3: Document retrieval [28] | Information retrieval to address questions patients may have when reading clinical reports. | Patients | Title: "thrombocytopenia treatment corticosteroids"<br>Description: "How long should be [sic] the corticosteroids treatment to cure thrombocytopenia?"* | On request |
| VQA-RAD [29] | Visual question-answering on radiology images using manually generated clinical questions. | Clinicians, radiologists | Q: "Is there evidence of an aortic aneurysm?"<br>A: "no" | Public |
| PathVQA [30] | Visual question-answering based on histopathological images from pathology textbooks and digital resources using automatically generated questions. | Pathologists | Q: “Is the covering mucosa ulcerated?”<br>A: “yes” | Public |
| IU X-Ray [31] | Radiological report generation | Radiologists | “Eventration of the right hemidiaphragm. No focal airspace consolidation. No pleural effusion or pneumothorax.”* | Public |
| PEIR Gross dataset [32] | Histological report generation | Pathologists | “Carcinoma: Micro low mag H&E needle biopsy with obvious carcinoma” | Public |
| PadChest [33] | Radiological report generation | Radiologists | “pleural effusion, costophrenic angle blunting” | On request |
| ImageCLEFmedical 2018 Visual Question Answering (VQA) [34] | Visual question-answering on clinical images extracted from PubMed using synthetic question-answer pairs based on image captions. | Clinicians | Q: “Is the lesion associated with a mass effect?”<br>A: “no” | Upon registration |
| ImageCLEFmedical 2019 Visual Question Answering (VQA) [34] | Visual question-answering on radiology images covering four question categories, i.e., modality, plane, organ system and abnormality. | Clinicians | Q: “What is the primary abnormality in this image?”<br>A: “Burst fracture” | Upon registration |
| ImageCLEFmedical Visual Question Answering (VQA) 2020 and 2021 [34] | Visual question-answering on radiology images focusing on questions about abnormalities. | Clinicians | Q: “What abnormality is seen in the image?”<br>A: “ollier’s disease, enchondromatosis” | Upon registration |
| ImageCLEFmedical Medical Automatic Image Annotation Task [34] | Annotation of radiology/histology images | Radiologists/Pathologists | “Immunohistochemical stain for pancytokeratin, highlighting tumor cells with unstained lymphocytes in the background” | Upon registration |

Mapping of AI benchmarks to the list of real-world clinical work activities revealed that many work activities with potential for assisting and disburdening healthcare staff are currently not or only scarcely addressed (Table 4). Work activities with the highest number of associated AI benchmarks were "*Process x-rays or other medical images*", "*Review professional literature to maintain professional knowledge*" and "*Gather medical information from patient histories*".

An informal review of the validity and representativeness of the currently available, directly clinically relevant benchmarks revealed two areas of concern: First, we found that datasets commonly contained redundant items, or items not representative of real-world clinical tasks. Examples of the latter include unrepresentative questions in visual question-answering tasks, such as *"What is not pictured in this image?"* or bogus synthetically generated questions, such as *"The tumor cells and whose nuclei are fairly uniform, giving a monotonous appearance?"*. Second, for manually or semi-automatically generated datasets, the varying expertise and number of annotators involved in the creation process might have impacted data quality.

**Table 4. Top 15 clinical work activities sorted by their desirability of automation reported by clinicians together with their current coverage by AI benchmarks.** Only work activities that could potentially be addressed by the types of NLP-centric AI covered in our analysis were included. The complete list of work activities is available from the online supplementary material (see Section 'Code and data availability'). *Directly relevant* benchmarks represent actual clinical tasks; *indirectly relevant* benchmarks represent low-level capabilities (e.g., 'clinical text classification') that can be utilized by clinical AI systems but do not represent clinical tasks themselves.

| **Clinical work activities** | **Task type** | **# Benchmarks** *directly relevant (indirectly relevant)* |
|---|---|---|
| Write reports or evaluations | Administrative | **3** (0) |
| Enter patient or treatment data into computers | Administrative | **0** (0) |
| Develop treatment plans for patients or clients | Clinical care | **0** (0) |
| Prepare official health documents or records | Administrative | **0** (0) |
| Verify accuracy of patient information | Clinical care | **0** (0) |
| Schedule patient procedures or appointments | Administrative | **0** (0) |
| Process medical billing information | Administrative | **0** (3) |
| Inform medical professionals regarding patient conditions and care | Clinical care | **0** (0) |
| Prepare scientific or technical reports or presentations | Scientific | **0** (0) |
| Verify that medical activities or operations meet standards | Clinical care | **0** (0) |
| Review professional literature to maintain professional knowledge | Clinical care | **4** (9) |
| Order medical supplies or equipment | Administrative | **0** (0) |
| Process healthcare paperwork | Administrative | **0** (0) |
| Gather medical information from patient histories | Clinical care | **1** (9) |
| Process x-rays or other medical images | Clinical care | **9** (24) |

# 4. Discussion

## 4.1. Principal results

In this paper, we have introduced the—to the best of our knowledge—first and most comprehensive catalogue of annotated NLP datasets and benchmark datasets relevant for clinical tasks. Based on this data source, we have analysed the overlap between tasks currently covered by benchmarks and tasks that healthcare personnel most want to see addressed.

We have created this dataset based on a systematic review conducted in accordance with current best practice guidelines, such as the PRISMA guidelines. [35] The PRISMA guidelines have primarily been created for systematic reviews of randomized trials and evaluation of interventions. Nonetheless, the general review framework is applicable to all types of systematic literature reviews.

We built our analysis of benchmark coverage of clinically relevant tasks on recent work by Frühwirt and Duckworth who investigated the possibility and desirability of automation of work activities in the healthcare domain based on thousands of ratings by domain experts. [22] We found that AI benchmarks of direct clinical relevance are scarce and fail to cover many work activities that clinicians most want to see addressed. Especially tasks associated with routine documentation and patient data administration workflows were scarcely represented despite significant associated workloads.

There are several potential reasons for this. First, our criteria for benchmark datasets were the availability of clear evaluation criteria and public accessibility of the dataset. Our analysis has shown that a large amount of published research on clinically relevant AI is conducted on datasets that are not available to other researchers. This is in line with previous research that investigated the availability of biomedical datasets. [16] Data governance in the biomedical domain and especially its clinical subdomains is strongly marked by privacy and data protection considerations. In the subfield of clinical applied AI research oftentimes institutional data, e.g. from the local EHR, is used. Making such data available to other researchers requires adequate measures to sustain patient privacy which may be associated with increased workloads and/or costs.

While the safeguarding of patient privacy and data protection is of fundamental importance, it also entails cutbacks in terms of transparency and reproducibility of current AI research in the biomedical domain which ultimately may negatively impact research progress. Emerging de-centralized learning approaches, such as federated learning, may address this problem by enabling the sharing of data even across institutions while maintaining patient privacy and data protection. [36] However, while such approaches hold promise to unlock the potential value of sensible data, widespread application is yet to be awaited and will require adequate incentivisation.

The lack of coverage of administrative clinical work activities may further point to a research prioritization of tasks that directly impact patient treatment, such as diagnosing a disease or finding information on diseases and/or their treatment. This seems to neglect that a significant amount of healthcare providers' workload is caused by administrative tasks and paperwork. [37,38] Disburdening healthcare providers from such administrative tasks may ultimately indirectly improve the quality of provided healthcare by freeing up time and cognitive resources for actual clinical care and patient communication.

We make the curated datasets available and expect it to be useful to a broad target audience, including biomedical and clinical researchers, NLP and AI researchers as well as ML practitioners in general. The dataset can be utilized in a variety of tasks such as application development, AI research as well as meta-research. For AI developers and researchers, the dataset offers a comprehensive overview of NLP in medical application areas and enables finding datasets and benchmarks relevant for a specific task and type of data (e.g., scientific, clinical). Its practical utility is further increased through the addition of detailed information on data accessibility and licensing.

One limitation of this initial version of the dataset is that we included datasets and benchmark datasets based on their occurrence in the literature and grey literature regardless of their size, provenance, generation/annotation process and internal validity. We therefore strongly encourage users to individually verify that the respective dataset is appropriate for the intended use case.

The dataset is intended as a living, extendable resource. Further, newly identified biomedical datasets will be added to the catalogue. To this end, methods for creating a semi-automatic pipeline for extracting datasets from the literature will be investigated. In addition, the dataset is open to additions and suggestions by users, which can be communicated directly in the Google sheet version of the dataset. The TSV version of the dataset will be versioned to allow comparability and tracking.

Finally, in this work, we have limited the scope to NLP tasks, including cross-domain tasks, such as visual question-answering. Other high impact clinical application domains of AI include computer vision tasks, such as classification of radiology or pathology images, or video-based tasks related to robotic and laparoscopic surgery. Conducting an analysis of benchmark datasets for these and other clinically relevant AI focus areas could be the subject of future research.

# 5. Conclusions

AI benchmarks of direct clinical relevance are scarce and fail to cover many work activities that clinicians most want to see addressed. Tasks associated with routine documentation and patient data administration workflows are seldom represented despite significant associated workloads.

Investing in the creation of high-quality, representative benchmarks for clinical tasks will have significant positive long-term impact on the utility of AI in clinical practice. Ideally, this should be addressed by allocating more funding to the development of such benchmarks, since their creation can be costly and is currently poorly incentivized.

# 6. Code and data availability

The data set is released in two formats: As a Google sheet[13] and as a versioned TSV file at Zenodo[14]. [23] Additionally, we make the raw exports of the literature review results available via Zenodo.

The Google sheet can be browsed online:
https://docs.google.com/spreadsheets/d/1QjUxxnZ3tuyW5dj6nkt_o5yJcWUZec4ttfJxO8Zlty4/

Hovering over the column names provides the user with a short description of the annotation fields. The datasets can be filtered using 'Filter views' available in the Google sheets menu.

[13] https://docs.google.com/spreadsheets/d/1QjUxxnZ3tuyW5dj6nkt_o5yJcWUZec4ttfJxO8Zlty4/
[14] DOI: 10.5281/zenodo.4647823

In addition, we have created three pre-customised filters to make the dataset more easily explorable.

Code to generate the summary statistics including additional statistics of the dataset are available from Github:
https://github.com/OpenBioLink/ITO/tree/master/notebooks/clinical_benchmarks

# 7. Acknowledgments

This work was supported by European Community's Horizon 2020 Programme grant number 668353 (U-PGx). The authors declare no conflicts of interest.

# 8. Declaration of interests

The authors declare no conflicts of interest.